 \documentclass[pdflatex,sn-basic,iicol]{sn-jnl} 

\usepackage{graphicx}%
\usepackage{multirow}%
\usepackage{amsmath,amssymb,amsfonts}%
\usepackage{amsthm}%
\usepackage{mathrsfs}%
\usepackage[title]{appendix}%
\usepackage{xcolor}%
\usepackage{textcomp}%
\usepackage{manyfoot}%
\usepackage{booktabs}%
\usepackage{algorithm}%
\usepackage{algorithmicx}%
\usepackage{algpseudocode}%
\usepackage{listings}%
\usepackage{color}
\usepackage{tikz}
\usepackage{array}
\usepackage{booktabs}

\theoremstyle{thmstyleone}%
\theoremstyle{thmstyletwo}%

\theoremstyle{thmstylethree}%

\begin{document}

\title[Article Title]{Visual Attention Graph}


\author*[1,2]{\fnm{Kai-Fu} \sur{Yang}}\email{yangkf@uestc.edu.cn}
\author*[1,2]{\fnm{Yong-Jie} \sur{Li}}\email{liyj@uestc.edu.cn}

\affil*[1]{\orgdiv{MOE Key Laboratory for NeuroInformation}, \orgname{School of Life Science and Technology, University of Electronic Science and Technology of China}, \orgaddress{\city{Chengdu}, \postcode{610054}, \state{Sichuan}, \country{China}}}
\affil[2]{\orgdiv{The Yangtze Delta Region Institute (Huzhou)}, \orgname{University of Electronic Science and Technology of China}, \orgaddress{\city{Huzhou}, \postcode{313001}, \state{Zhejiang}, \country{China}}}


\abstract{Visual attention plays a critical role when our visual system executes active visual tasks by interacting with the physical scene. However, how to encode the visual object relationship in the psychological world of our brain deserves to be explored. In the field of computer vision, predicting visual fixations or scanpaths is a usual way to explore the visual attention and behaviors of human observers when viewing a scene. Most existing methods encode visual attention using individual fixations or scanpaths based on the raw gaze shift data collected from human observers. This may not capture the common attention pattern well, because without considering the semantic information of the viewed scene, raw gaze shift data alone contain high inter- and intra-observer variability. To address this issue, we propose a new attention representation, called \textit{Attention Graph}, to simultaneously code the visual saliency and scanpath in a graph-based representation and better reveal the common attention behavior of human observers. In the attention graph, the semantic-based scanpath is defined by the path on the graph, while saliency of objects can be obtained by computing fixation density on each node. Systemic experiments demonstrate that the proposed attention graph combined with our new evaluation metrics provides a better benchmark for evaluating attention prediction methods. Meanwhile, extra experiments demonstrate the promising potentials of the proposed attention graph in assessing human cognitive states, such as autism spectrum disorder screening and age classification. }




\maketitle

\section{Introduction}
Visual attention plays a critical role in the active visual tasks by performing interaction between human cognitive system and the external scenes, while eye movements specifically manifest this interactive processing. Predicting where people look and how they scan a visual scene can help us understand the cognitive processes behind visual attention \citep{eckstein2011visual,wolfe2021guided}. Moreover, developing computational models of visual attention and scanpath prediction can also contribute to improve computer vision algorithms \citep{nguyen2018attentive}.

The saliency detection task is well-defined to yield a saliency map indicating regions that attract human attention in natural scenes under free-viewing or task-driven conditions. A substantial number of studies on saliency detection have been developed since the pioneering work by Koch and Itti \citep{koch1987shifts, itti1998model}, and extensively reviewed in literature \citep{borji2012state, borji2019saliency}. The research of saliency models is prosperous with well-established benchmarks \citep{borji2013analysis}, in which the distribution of human fixations recorded with eye trackers (e.g., fixation density obtained by smoothing human fixation points with a Gaussian kernel) is usually used as groundtruth data to evaluate the models of saliency detection. However, these saliency detection methods only focus on the spatial distribution of fixations, ignoring the temporal dynamics of attention shifts.

\begin{figure}[tb]
\centering
\includegraphics[width=3.0in]{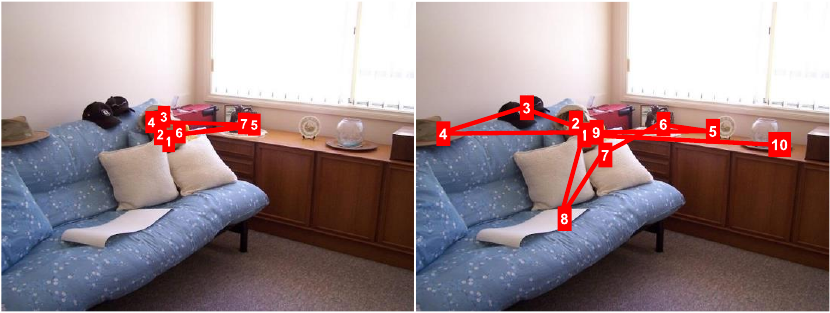}
\caption{Examples show the high intra-observer variability of scanpaths that are from two different human observers when viewing the same scene. Noted that the data is provided by the OSIE dataset \citep{xu2014predicting}.}
\label{FigScanExp}
\end{figure}

In contrast, scanpath prediction tries to take the temporal dimension of the gaze into account and aims to predict the path of gaze shifting. Recently, scanpath prediction is an emerging topic that has attracted increasing interest and many scanpath prediction models have been proposed \citep{kummerer2021state}. Currently, scanpath prediction models are mainly evaluated using the raw gaze-shift data from human observers, which always contain high inter- and intra-observer variability \citep{le2016introducing}. This means that human gaze-shifts differ significantly among and within individuals even when viewing same scenes, and thus a single or few observers’ data may not represent the common patterns of human visual attention well. Therefore, comparing the predicted scanpath with individual gaze-shift data may not be a suitable way for evaluating scanpath prediction models. 

To address these challenges, it is expected to build novel representation and evaluation methods of eye-movement-based visual attention, aiming to leave out inter- and intra-observer variability. To begin with, let us first clarify several specific issues, which are also the main motivations of this study.

\textit{\textbf{(1) Besides saliency map and scanpath, why do we need a new representation of visual attention?}}  
There is a reasonable assumption that observers’ gaze-shift data can reveal common patterns of visual attention, despite the inter- and intra-observer variability \citep{ellis1986statistical, martin2022probabilistic}.  Therefore, visual attention representation should indicate these common behavioral patterns, which can express the cognitive processes behind attention when we view and understand scenes. As mentioned, widely used attention representation includes saliency map and scanpath. Saliency maps represent the spatial distribution of fixations, but ignore the temporal dynamics of attention shifts. In contrast, individual scanpaths contain gaze shifting, but show high inter- and intra-observer variability. Fig. \ref{FigScanExp} shows an example of intra-observer variability of scanpaths when freely viewing the same scene. This motivates us to explore new representations of common attention patterns from the gaze-shift data containing unavoidable variability.
 
\begin{figure*}[tb]
\centering
\includegraphics[width=3.5in]{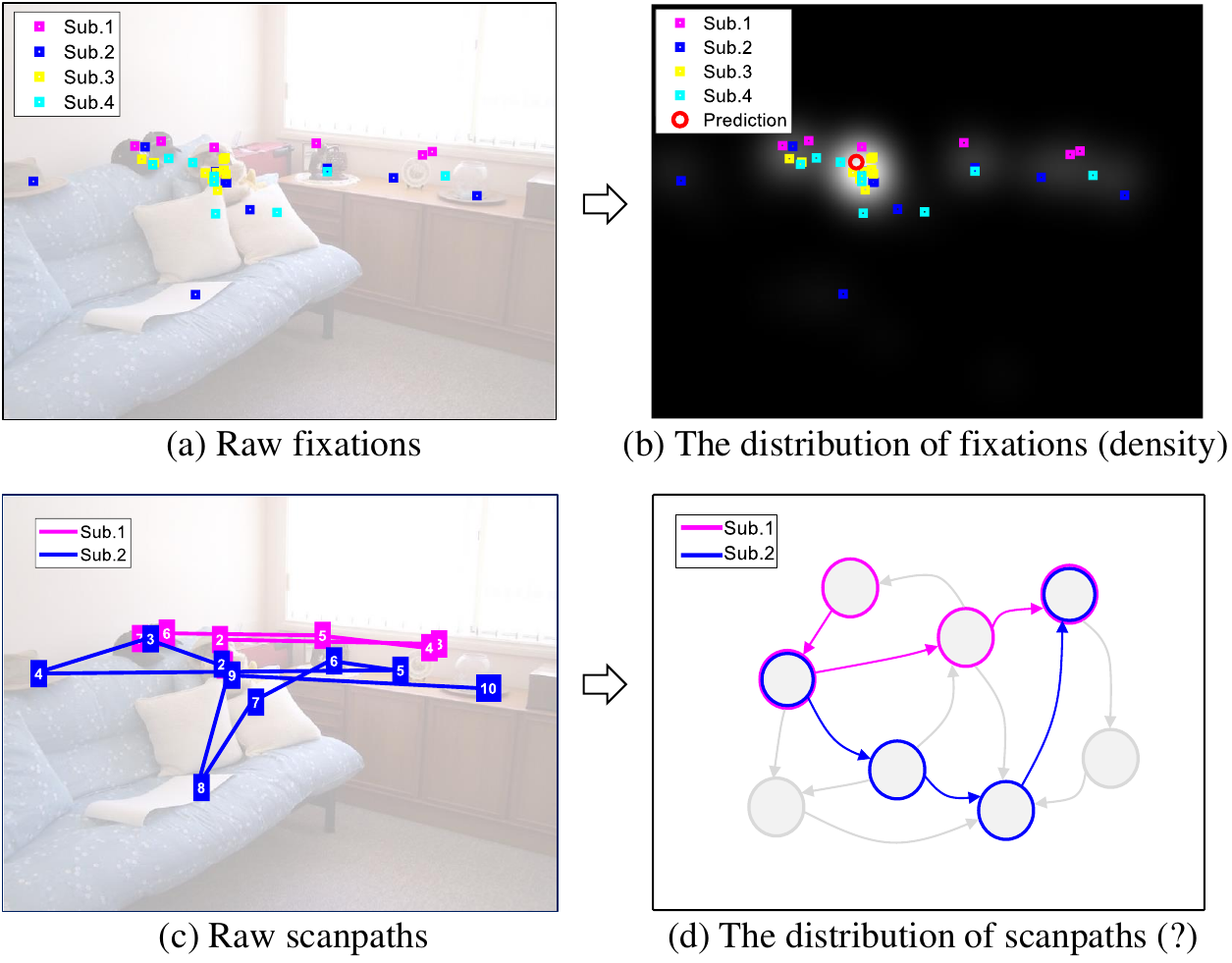}
\caption{Comparison between different attention-related representations. (a)-(b) illustration of the ideas of evaluating fixation prediction task, (c)-(d) illustrating  the requirement of distribution of human scanpaths to evaluate scanpath prediction models, referring to the evaluation of fixation prediction task. }
\label{FigScanDis}
\end{figure*}

\textit{\textbf{(2) What is a good representation of visual attention?}}   
Previous studies support that visual attention is semantic-based \citep{sun2003object, xu2014predicting, moore1998object, peng2022contour, zhang2022scanpath} and can naturally represent the hierarchical selectivity of attentional shifts \citep{sun2003object}.  From Fig. \ref{FigScanExp} we can also find that most fixations are located on a few semantic objects. Therefore, we argue that a new object-based representation should code the common patterns of visual attention and semantic-based scanpath is excepted to leave out inter- and intra-observer variability. In addition, single or few semantic scanpaths are difficult to capture the distribution of human scanpaths. To clarify this point, let’s refer to the evaluation of fixation prediction models. Fixation density is a good representation of the distribution of human fixations reconstructed from the data of several observers. For example, Fig. \ref{FigScanDis}(a) shows fixations from four observers that are marked in squares with different colors, and the intensity in Fig. \ref{FigScanDis}(b) indicates the distribution of human fixations reconstructed from these four observers’ data. Thus, one predicted fixation (red circle) that is obviously different (not overlap) from the sampled human fixations can be considered a good estimation because it is located in the high-confidence region of fixation density. Similarly, a good representation of the distribution of human scanpaths (Fig. \ref{FigScanDis}(d)) is also expected by reconstructing from sampled human scanpaths shown in Fig. \ref{FigScanDis}(c). In summary, a good representation of visual attention should be semantic-based to reduce inter- and intra-observer variability and express the distribution of visual scanpaths.

\textit{\textbf{(3) How to build and evaluate the representation of visual attention?}}
This is the main concern of this study. Our goal is to construct a new representation of visual attention that goes beyond raw fixation and gaze-shift data to reveal the common patterns of visual attention while leaving out inter- and intra-observer variability. We first build semantic scanpaths from the raw scanpaths by grouping fixations in  same semantic terms (e.g., objects) and then construct a graph-based representation to describe the distribution of semantic scanpaths. Meanwhile, based on this newly-defined attention representation and evaluation metrics, we further explore the potential contribution of our attention representation in real word applications.

The main contributions of this study can be summarized as follows:
\begin{itemize}
\item{A graph-based representation of attention (called \textit{Attention Graph, i.e., AG}) is proposed based on the definition of semantic scanpath, which can better reveal the intrinsic common patterns of visual fixations and gaze shift behaviors of human observers when viewing a visual scene.}
\item{A corresponding benchmark is built for evaluating the methods of semantic scanpath prediction, including a graph-based groundtruth to describe the distribution of scanpaths of observers and new metrics for quantitative evaluation.}
\item{Numerous existing scanpath prediction methods are evaluated on the proposed benchmark based on the attention graph. Additionally, the attention graph is applied in the tasks of autism spectrum disorder screening and age classification to demonstrate the potentials of the proposed method.}
\end{itemize}

\section{Related Works}
\subsection{Saliency Detection}
The saliency map was first suggested by Koch et al. to represent visual spatial attention in the early visual system \citep{koch1987shifts}. Classical saliency detection methods are typically based on the feature integration theory, which posits that visual attention is driven by low-level visual cues such as local luminance, color, and orientation \citep{treisman1980feature,itti1998model, itti2001computational}. These methods draw inspiration from the hierarchical and parallel visual pathways in early vision, suggesting that local regions with high contrasts in scenes attract more attention in a stimulus-driven manner.

In addition, perceptual cues such as the well-known Gestalt principles \citep{Koffka1935Principles}  are also crucial factors for visual attention \citep{yu2016computational, peng2021saliency}. For instance,  the closure and symmetry of regions are used to facilitate saliency detection \citep{kootstra2008paying, kootstra2010using, peng2022contour, yang2016unified}.  Guided search theory provides a more general framework for modeling visual attention \citep{wolfe2021guided}.  Beyond local contrast, scene context and global information play more important roles in guiding visual attention \citep{torralba2006contextual}. Existing studies have demonstrated that typical structural cues (such as layout, openness, and depth) can improve visual saliency detection \citep{borji2016vanishing, wolfe2017visual}.

Research on saliency detection has flourished over the last few decades, resulting in numerous well-established benchmarks \footnote{https://saliency.tuebingen.ai/}. For instance, Vig et al. made the first attempt to predict human fixation using convolutional neural networks (CNNs) \citep{vig2014large}. Then, DeepGaze series employed deeper CNNs for saliency prediction and achieved state-of-the-art performance on certain datasets \citep{Kummerer2014deep, Kummerer2016DeepGaze, kummerer2022deepgaze, linardos2021deepgaze}. Additionally, various CNN variants were designed to improve saliency detection, including approaches that combine multi-level features \citep{Cornia2016Deep, wang2017deep} and fully convolutional neural networks \citep{Kruthiventi2017DeepFix}. Saliency detection methods on single images have achieved  promising performance by introducing deep learning technology \citep{borji2019saliency}. Recently,  research in saliency detection has expanded to novel data domains, such as light field saliency detection \citep{gao2023thorough}, video saliency detection \citep{guo2017video}, and RGB-D saliency detection \citep{li2023robust, wang2023uniform}.  

\subsection{Scanpath Prediction}
To develop and evaluate scanpath prediction models, multiple datasets have been collected with eye-tracking experiments in the context of free-viewing and visual search tasks. The task of scanpath prediction with free-viewing condition usually shares the datasets collected for the fixation prediction task.  For example, MIT1003 \citep{judd2009learning}, CAT2000 \citep{borji2015cat2000}, and OSIE \citep{xu2014predicting} are widely used to explore scanpath prediction task in previous works \citep{kummerer2021state}. Specifically, the OSIE \citep{xu2014predicting} dataset contains 700 images with eye-tracking data of 15 observers and annotation data of segmented objects and 12 semantic attributes (e.g., smell, taste, etc.), which are useful for analyzing the semantic-level information in visual attention. On the other hand, some studies have focused on predicting goal-directed attention in visual search tasks, for which several datasets were provided, such as COCO-search18 \citep{chen2021coco} and the dataset provided by Koehler et al. \citep{koehler2014saliency}.

Numerous studies focus on predicting human scanpaths \citep{kummerer2021state}. Early methods mainly draw inspiration from the biological visual system. For instance, Itti et al. built a method to generate scanpath from the saliency map with a biologically inspired inhibition-of-return (IOR) and winner-takes-all (WTA) mechanisms \citep{itti1998model}. Wang et al. proposed the saccade model by integrating reference sensory responses, fovea-periphery resolution discrepancy, and visual working memory \citep{wang2011simulating}. Other methods predict scanpath based on certain statistical properties of human scanpaths. For example, some authors modulated the jump distribution for scanpath prediction \citep{coutrot2018scanpath, le2016introducing}, while others modelled scanpaths with statistical learning methods \citep{liu2013semantically, xia2019predicting, jiang2016learning}. Recently, deep-learning methods have also been widely used for the task of scanpath prediction, such as SaltiNet \citep{assens2017saltinet}, IOR-ROI-LSTM \citep{sun2019visual}, PathGAN \citep{assens2018pathgan}, and DeepGaze III \citep{kummerer2019deepgaze}. 

Compared to many works that use bottom-up information to guide visual attention from one location to the next \citep{koch1987shifts, itti1998model}, object-based attention theory argues that attention is actually directed to an object or a group of objects \citep{duncan1984selective, scholl2001objects}. One more related work is the object-based visual attention model proposed by Sun and Fisher \citep{sun2003object}, which provides a hierarchical object-based visual attention system. By contrast, this paper further develops a new attention graph representation based on the object-based scanpath and builds the corresponding benchmark to evaluate the models of semantic scanpath prediction.  

\begin{figure*}[tb]
\centering
\includegraphics[width=6.0in]{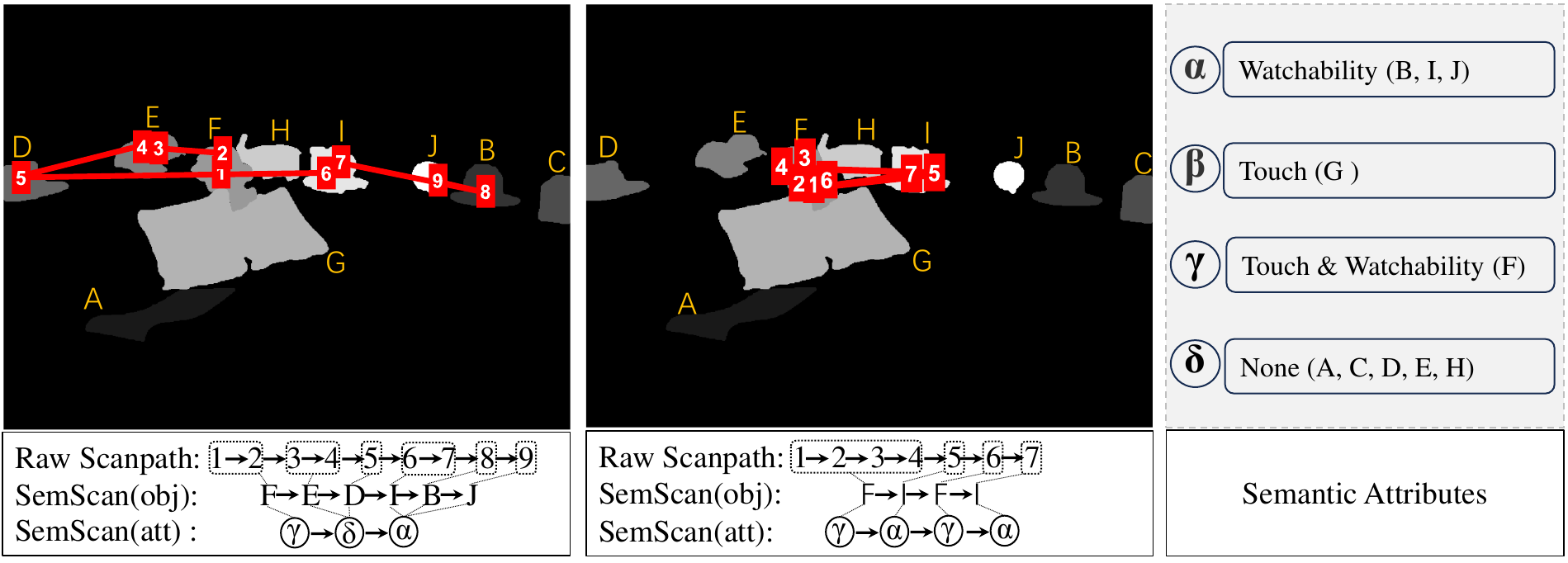}
\caption{Examples for demonstrating the generative process of semantic scanpath from raw fixations of two observers, and the annotations and attributes of objects, which are provided by the OSIE dataset \citep{xu2014predicting}. Specifically,  the object masks are used to group raw fixations into SemScan(obj), while 12 semantic-level attributes (Smell, Touch, Watchability, etc.) provided in the OSIE dataset are used to further group SemScan(obj) into SemScan(att). It should be noted that one object with multiple attributes (e.g., Touch \& Watchability in this figure) is treated as a new semantic attribute. Meanwhile, \textit{None} indicates the objects without labeled attributes.}
\label{FigDefSS}
\end{figure*}

In terms of evaluating scanpath prediction models, researchers have employed various metrics \citep{fahimi2021metrics}. For instance, Wang et al. used the time-delay embedding method to evaluate scanpath prediction \citep{wang2011simulating}. Some of the more commonly used string-based metrics include ScanMatch \citep{cristino2010scanmatch} and Sequence Score \citep{borji2013analysis}. The ScanMatch method encodes fixation sequences into strings and computes the similarity score of two fixation sequences using the Needleman-Wunsch algorithm \citep{needleman1970general}. In the encoding step, fixation locations are quantized into spatially and temporally bins according to gridded areas to retain fixation locations. Subsequently, Borji et al. proposed Sequence Score to improve ScanMatch by dividing each scene with clustering fixations instead of gridded areas\citep{borji2013analysis}. The Sequence Score is more fairly for evaluating the performance of scanpath models, as clustering fixations can leave out intra-observer variability to some certain. However, clustering fixations only rely on spatial distance is still questionable. For example, some fixations gathering near the junction of two objects are treated as one term of scanpath, although they fall into two different semantic regions (objects). Basically, almost all current scanpath models are evaluated by comparing them with the individual gaze-shift data (as individual groundtruth scanpath) and then taking the mean metric over observers. We argue that existing evaluation methods cannot well capture the distribution of human scanpaths. 

Recently, Xia et al. reviewed the existing metrics for saccade prediction and proposed a new method for evaluating scanpath models using a recurrent neural network-based metric \citep{xia2020evaluation}. The learned metric showed good consistency with human assessment in measuring scanpath similarity. While deep neural networks provide a strong capability to predict similarities between two scanpaths, it is difficult to explain why they are similar. Explainability is important, especially in scanpath prediction, because one of the important goals of scanpath prediction is to explain visual attention behavior. 

\subsection{Application of Visual Attention}
Generally, saliency detection plays a crucial role in subsequent computer vision tasks, such as image compression \citep{christopoulos2000jpeg2000}, image re-targeting \citep{goferman2011context}, and object recognition \citep{rutishauser2004bottom}. Recently,  the concept of visual attention has become fundamental in developing deep neural networks. Within deep learning systems, attention mechanisms modulate information flow and feature fusion, leading to improved feature learning performance.  Common categories of attention used in deep neural networks include channel attention, spatial attention, and temporal attention \citep{guo2022attention}. Numerous studies have demonstrated the benefits of attention in various tasks, including image recognition \citep{hu2018squeeze}, object detection \citep{dai2017deformable}, and segmentation \citep{fu2019dual}.  Notably, the self-attention and its variants have directly contributed to develop transformer-based methods that had a great success in the field of natural language processing \citep{vaswani2017attention} and computer vision \citep{dosovitskiy2020image}. 

In the field of medicine, the individual’s viewing behavior is closely linked to higher-order cognitive processes. Consequently, viewing behaviors such as scnapaths and fixations serve as efficient cues for screening some neurodevelopmental disorders, such as autism spectrum disorder (ASD) \citep{wang2015atypical} and ADHD \citep{deng2022detection}. Recent studies have demonstrated the high accuracy in ASD diagnosis  based on visual viewing patterns \citep{chen2019attention,xia2022dynamic, jones2023development}.  Additionally, eye tracking provides insights into how radiologists evaluate and diagnose medical images \citep{wolfe2021experts,wolfe2022eye}. Integrating doctor's attention into computer-aided diagnostic systems has been shown to effectively improve the efficiency and interpretability of computational methods \citep{wang2022follow,zhao2024mining}. These methods are designed mainly based on the spatial distribution of visual attention but with insufficient emphasis on the temporal dynamics of attention shifts, i.e., scanpath.

\section{Method}
In this section, we introduce a novel semantic-based representation of visual attention, called \textit{Attention Graph}. The goal of the attention graph is to encode both visual attention and semantic scanpaths in a graph-based representation, aiming to better reveal the common attention behaviors by leaving out intra-observer variability of human gaze shifts. Additionally, new metrics based on attention graph are designed to comprehensively evaluate the semantic scanpath prediction and attention graph applications.

\subsection{Definition of Attention Graph}
\subsubsection{Semantic Scanpath}
\label{sec.ssdef} 
First, we define the semantic scanpath by grouping the fixations in the same semantic regions (e.g., objects) into a single scanpath term. We construct the semantic scanpath based on the raw fixations collected from human observers, and the annotations and attributes of objects. Previous studies suggest that visual attention is object-based \citep{duncan1984selective, scholl2001objects, sun2003object, zhang2022scanpath}. Therefore, we first group raw fixations based on object annotation in order to generate a new object-based representation of human scanpath. Specifically, fixations that are temporally adjacent and spatially located in the same object regions are grouped into single elements of the object-based scanpath. 

\begin{figure}[tb]
\centering
\includegraphics[width=3.0in]{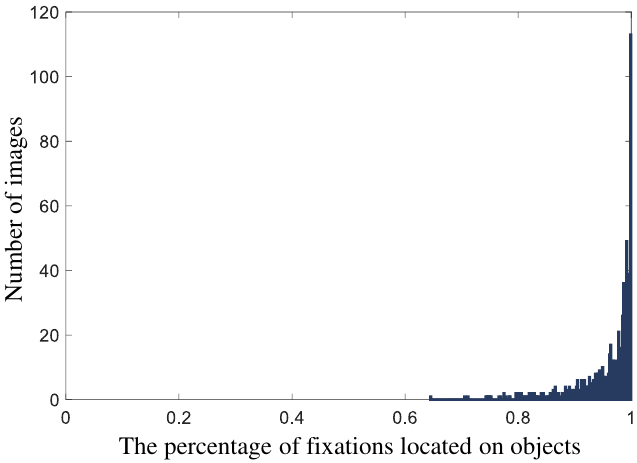}
\caption{Statistical analysis of fixations located inner objects in the OSIE dataset. }
\label{FigFixObj}
\end{figure}

Fig. \ref{FigDefSS} illustrates two examples that clarify the generation process of the semantic scanpath from raw fixations and object annotations from the OSIE dataset \citep{xu2014predicting}. Specifically, along with the fixation sequence, temporally adjacent fixations are first encoded in a single item once they are located within the same object to obtain semantic scanpath. From Fig. \ref{FigDefSS}, we can also observe that the semantic scanpath retains the review shifts because they also reflect a kind of cognition processing. It should be noted that we remove fixations located outside all objects at least beyond 30 pixels. This is reasonable considering that some fixations could be located near an object that is paid attention to, due our peripheral vision. We also conduct a simple statistical analysis on the OSIE dataset, which indicates that more than 95\% (95.5\%) of fixations are located in or near-annotated objects. Fig. \ref{FigFixObj} shows that most fixations are located inner objects for almost all images in the OSIE dataset.

Furthermore, we use the term of semantic scanpath considering that scanpaths can be represented at varying semantic levels. On one hand, objects may consist of multiple parts that can individually attract human attention. Therefore, segmenting the visual scene into local  parts or regions can build region-based scanpath with the proposed method. On the other hand, multiple objects might also share a common high-level semantic attribute. Therefore, objects with shared semantic attributes can be further grouped into a single element to generate an attribute-based scanpath (shown in Fig. \ref{FigDefSS}). By adopting these definitions, the semantic scanpath actually captures attention shifts across different semantic levels, offering a hierarchical representation of visual attention and gaze shifting.

In our subsequent experiments, we denote the object-based scanpath as \textit{SemScan(obj)} and the attribute-based scanpath as \textit{SemScan(att)}. Specifically,  12 semantic-level attributes (Smell, Touch, Watchability, etc.) provided in the OSIE dataset \citep{xu2014predicting} are used to further group SemScan(obj) into SemScan(att) in this study. We do not investigate part or region-based scanpaths because they are similar with that defined by ScanMatch \citep{cristino2010scanmatch} and Sequence Score \citep{borji2013analysis} methods which are not dedicated to capturing the semantic information. 

\begin{figure*}[tb]
\centering
\includegraphics[width=6.0in]{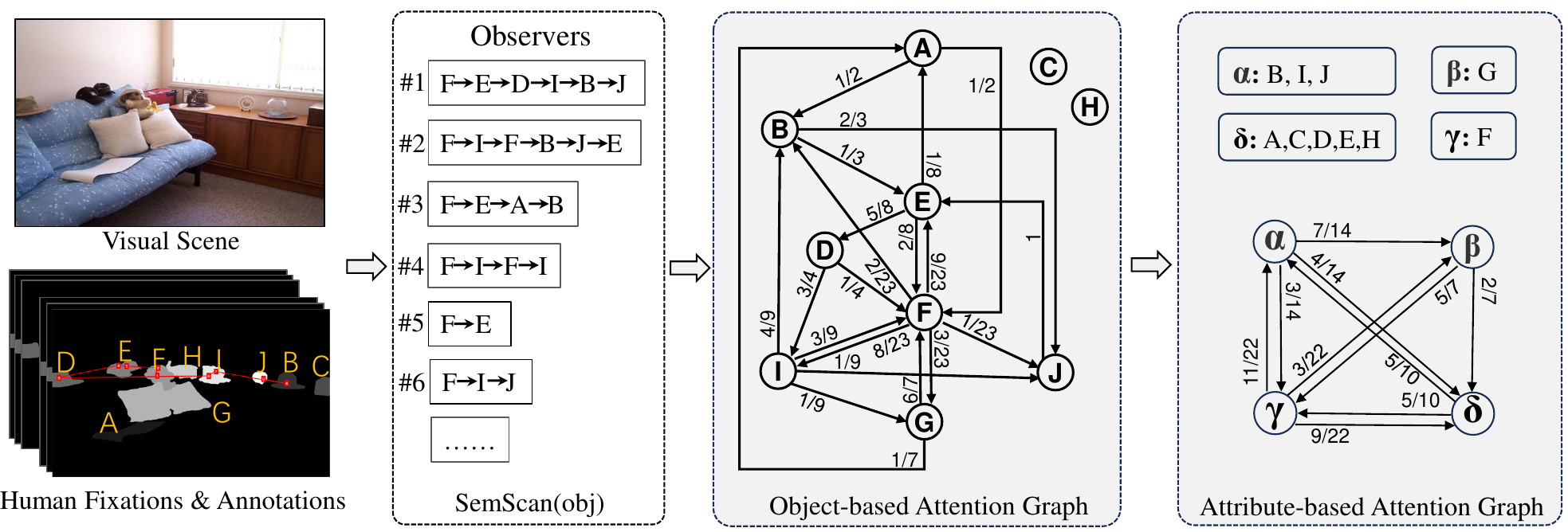}
\caption{Constructing the attention graph from multiple semantic scanpaths of observers. For example, the  weights from object \textit{\textbf{I}} to object \textit{\textbf{B}} is $4/9$ in the attention graph, which indicates that 4 out of 9 attention shifts from object \textit{\textbf{I}} to object \textit{\textbf{B}} in all observers are counted. Note that the edge weights are shown as fractional number to indicate raw counts of attention shifts. }
\label{FigAttGraph}
\end{figure*}

\begin{figure*}[tb]
\centering
\includegraphics[width=6.0in]{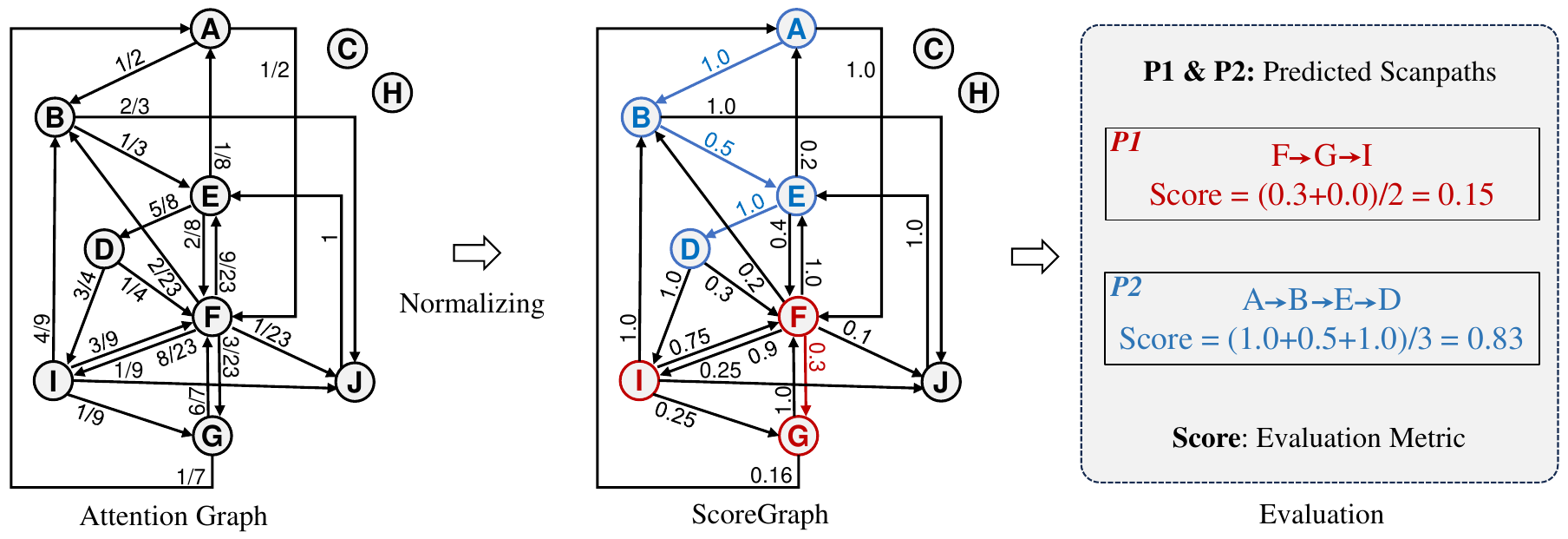}
\caption{The computational flow for evaluating predicted semantic scanpaths. }
\label{FigScore}
\end{figure*}

\subsubsection{Attention Graph}
For each scene, the fixations of each observer can be represented as an individual semantic scanpath. As shown in Fig. \ref{FigScanDis}, it is important to build a representation of the distribution of semantic scanpaths for the description of human attention. In this work, we propose a graph-based representation (i.e., \textit{Attention Graph}) that describes the overall distribution of all observers’ gaze shifts on a scene.

Formally, given an image containing a set of object annotations and semantic scanpaths from multiple observers, we define the attention graph as a weighted directed graph $\mathbb{G}$, i.e., $\mathbb{G}=(\mathbb{O},\mathbb{E})$, where $\mathbb{O}$ is a set of nodes and $\mathbb{E}$ is a set of edges. The nodes represent the objects present in the scene, and the weights of edges are allocated by the probability of attention shifts of all semantic scanpaths over observers between two objects.

An example of building attention graph for a scene is shown in Fig. \ref{FigAttGraph}. Specifically, the object-based scanpaths are first generated from all observers' fixations following the semantic scanpath definition in Section \ref{sec.ssdef}. All object-based scanpaths are combined into a weighted directed graph by computing shift probabilities of observers to build the object-based attention graph. According to this definition, the attention graph can be considered as a distribution of human scanpaths fitted from several observers' fixation data. This processing is similar to reconstructing fixation density as the distribution of human fixations from observers’ data. For example, it is reasonable to generate a new scanpath by sampling from the attention graph according to the distribution of weights. We believe that it is also an efficient representation of groundtruth for the evaluation of scanpath prediction. 

Naturally,  defining the semantic scanpath at different semantic levels will generate distinct representations of attention graphs. We propose that attention graphs can also be represented at varying semantic levels to create hierarchical attention graphs. For example, objects play a crucial role in visual perception, particularly given the widely-accepted concept of object-level visual attention \citep{sun2003object}. Consequently, our focus in this work is on object-based attention graphs. Additionally, exploring higher-level semantic attributes (such as smell, touch, and watchability) can help evaluate the utility of such semantic representations \citep{xu2014predicting}. For instance, the nodes (objects) of object-based attention graph with the same attribute are further grouped into single items, and the attribute-based attention graph is obtained (shown in Fig. \ref{FigAttGraph}). 

\subsection{Metrics Based on Attention Graph }
\label{3.3}
Based on the attention graph representation, new metrics are designed to evaluate the scanpath prediction models by comparing a predicted semantic scanpath with the attention graph. Specifically, the attention graph is first normalized for each node individually to obtain a \textit{ScoreGraph}. For example, the weights of all edges starting from node $A$ are divided by the maximum weights, ensuring that all weights are distributed within the range of $[0,1]$. Thus, each edge in the ScoreGraph can be treated as a score when the corresponding gaze shift happens. Finally, the score of one predicted scanpath is defined as the average score along the path of predicted scanpath in the ScoreGraph. 

\begin{figure*}[tb]
\centering  
\includegraphics[width=6.0in]{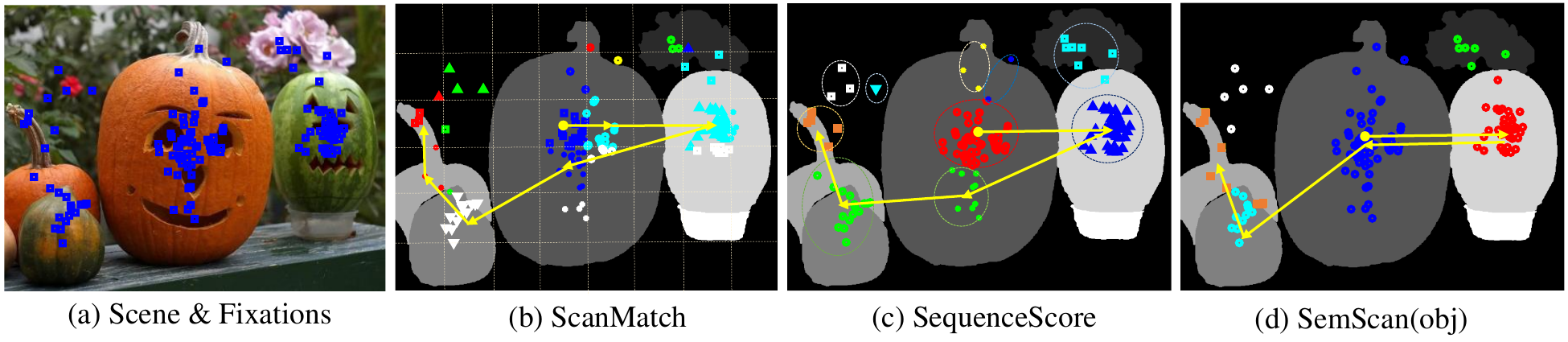}
\caption{Comparisions of different ways to code scanpath in ScanMatch\citep{cristino2010scanmatch}, Sequence Score\citep{borji2013analysis}, and the proposed SemScan(obj).}
\label{FigComScan}
\end{figure*}

Formally, given a predicted semantic scanpath $P=(o_1, o_2,…,o_k)$, and the length of sequence is ($K-1$). $o_k$ represents the $k^{th}$ term of the semantic scanpath.  In addition, $W_{(A \to B)}$ indicates the score of gaze shift from $A$ to $B$ in the ScoreGraph. The score of the predicted semantic scanpath can be obtained as 

\begin{equation} 
\label{E1} 
S_{scan} = \frac{1}{{K - 1}}\sum\limits_{t = 1}^{K - 1} {{W_{({o_t} \to {o_{t + 1}})}}} 
\end{equation}

Fig. \ref{FigScore} shows the computational flow for evaluating  predicted semantic scanpaths. For example, one predicted semantic scanpath $P1$ is $F \to G \to I$  (marked in red in Fig. \ref{FigScore}), the scores of $F \to G$  and $G \to I$ are 0.3 and 0.0, respectively. Note that the score of the missing connection is set to 0.0. Thus, the score of  $F \to G \to I$ is $(0.3+0.0)/2$, i.e., 0.15. This is a low score because the $P1$ contains a false alarm prediction of $G \to I$ compared to the human scanpaths.  In contrast,  another predicted semantic scanpath $P2$ is $ A \to B \to E \to D$  (marked in blue in Fig. \ref{FigScore}), which produces a higher score of 0.83.

\begin{figure*}[tb]
\centering
\includegraphics[width=6.0in]{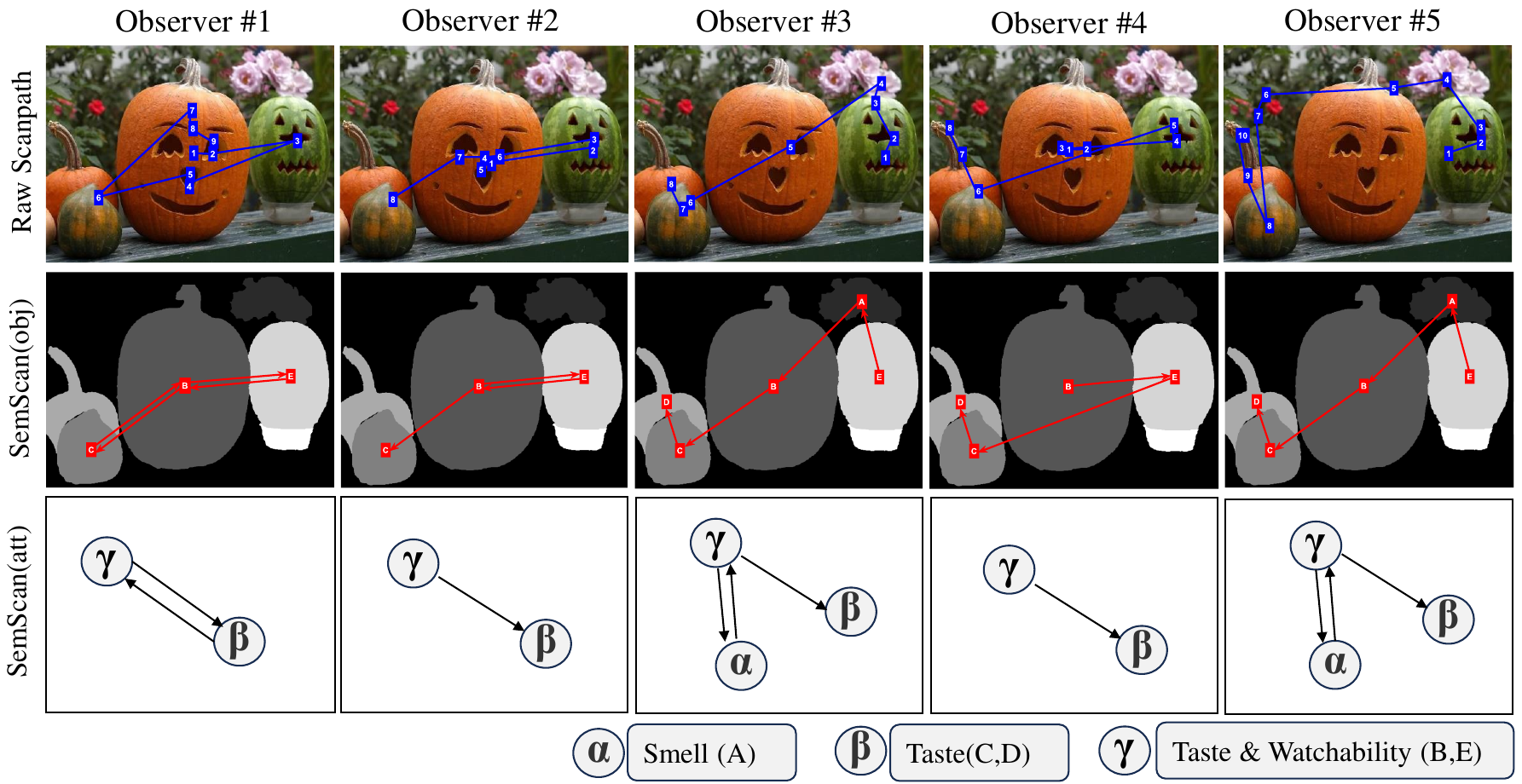}
\caption{Comparisons of the raw scanpath, the SemScan(obj), and the SemScan(att) of five observers.  Note that \textit{Smell}, \textit{Touch},\textit{ Taste \& Watchability} indicate the semantic-level attributes of each object  provided by the OSIE dataset \citep{xu2014predicting}. }
\label{FigHeScan}
\end{figure*}

It can be observed that the score obtained with Eq. (\ref{E1}) purely reflects the consistency of gaze shifts, regardless of saliency of object on each node. Therefore, an alternative metric that combines the evaluation of semantic scanpath and saliency of each nodes should be more reasonable for semantic scanpath evaluation. The saliency of an object can be obtained by summarizing fixation density located in it. When denoting the saliency of objects in a given predicted semantic scanpath $P=(o_1, o_2, …, o_k)$ as $\lambda(o_t) , o_t \in \{o_1, o_2, …, o_k\}$, the saliency-weighted score ($S'_{scan}$) can be defined as 

\begin{equation} 
\label{E2} 
S'_{scan} = \frac{1}{\mu}\sum\limits_{t = 1}^{K - 1} {\lambda(o_t) \cdot {W_{({o_t} \to {o_{t + 1}})}}} 
\end{equation}
where $\mu=  \sum\nolimits_{t = 1}^{K - 1} {\lambda (o_t)}, o_t \in \{o_1, o_2,…,o_k\}$.

\section{Experimental Results}
In this section, we first analyze the proposed semantic scanpath by comparing it with the widely-used coding of scanpath. We then evaluate multiple popular scanpath prediction methods based on our attention graph. In addition, we demonstrate the applications of the attention graph in evaluating cognition-related tasks, including toddlers' age classification and ASD screening. 

\subsection{Semantic Scanpath and Attention Graph} 
\label{sec.eval}
ScanMatch \citep{cristino2010scanmatch} and Sequence Score \citep{borji2013analysis} are two widely-used metrics to evaluate scanpath prediciton methods. From the definitions of these metrics, we find that they have actually left out intra-observer variability by gridding or clustering fixations into scanpath sequences based on the spatial distribution of fixations. Therefore, we first analyze and show the different ways to encode the raw fixations into scanpaths in these methods.

Fig. \ref{FigComScan} compares our SemScan(obj) with the coding of scanpath in ScanMatch\citep{cristino2010scanmatch} and Sequence Score \citep{borji2013analysis}. The ScanMatch (Fig. \ref{FigComScan}(b)) creates a saccadic sequence based on the spatially and temporally bins, and hence the scanpath is strongly dependent on the setting of number of bins \citep{cristino2010scanmatch}. Differently, the Sequence Score (Fig. \ref{FigComScan}(c)) generates saccadic sequence by clustering the fixations \citep{borji2013analysis}. Thus, the fixations on single large object might be divided into multiple clusters, even though they are located in the same semantic object. In contrast, Fig. \ref{FigComScan}(d) shows the result of the proposed SemScan(obj), in which all fixations located in the same objects are grouped into single items of scanpath. We believe it is reasonable to assume that the fixations in the same object indicate same semantic information when the observer views the scene.

Furthermore, to encode the scanpath in different semantic levels, we further build SemScan(att) by grouping objects with the same semantic attribute. Fig. \ref{FigHeScan} shows comparisons of the raw scanpath, the SemScan(obj), and the SemScan(att). From Fig. \ref{FigHeScan}, we can find that intra-observer variability decreases with the semantic coding. For example, raw scanpaths of different observers show large variability in both locations and orders of fixations. In contrast, the SemScan(obj) describes attention shifts among several main semantic objects and decreases the intra-observer variability of exact locations of fixations. In SemScan(att), the attention shifts happen among more abstract semantic attributes and all observers perform more consistent gaze shifting, e.g., from $\gamma$ to $\beta$.

\begin{figure*}[tb]
\centering
\includegraphics[width=6.0in]{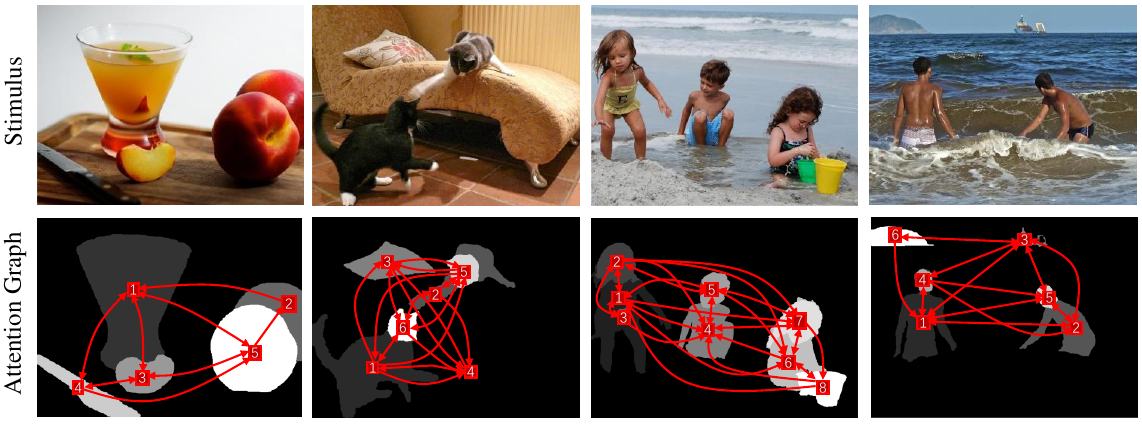}
\caption{Examples of human attention graphs that are built based on the fixation data and object annotations. Edge weights are ignored for the convenience of showing the graph structure.}
\label{FigAGcom}
\end{figure*}

Fig. \ref{FigAGcom} presents several examples of the human attention graph. Attention graph expresses the distribution of gaze shifting among semantic objects in the visual scene, which encodes the visual object relationship in the psychological world beyond physical scene. For example, from the attention graph listed in the third column of Fig. \ref{FigAGcom}, we can find that the observers first perceive the persons and then focus on their faces. This may reflect the cognitive strategy of coarse-to-fine when understanding visual scene \citep{hegde2008time}. Therefore, we can explore the cognitive strategy in active vision based on our attention graph representation. 

In addition, considering that the attention graph represents the distribution of semantic scanpaths, we can obtain human-like semantic scanpaths by sampling paths between any two given nodes from the attention graph. Thus, the proposed attention representation suggests a potential way to generate large scale visual scanpaths, which could contribute to augment training data when building deep neural networks for attention-related tasks, e.g., scanpath prediction.

\subsection{Evaluation of Scanpath on Attention Graph} 
Numerous of scanpath prediction methods are evaluated using the proposed attention graph representation. Firstly, classical methods usually generate scanpaths from the saliency map with the WTA operator \citep{itti1998model}. Recently, deep learning methods have also been used for scanpath prediction, such as SaltiNet \citep{assens2017saltinet}, PathGAN \citep{assens2018pathgan}, and VQA \citep{chen2021predicting}. In addition, two baselines are considered for comparison. The \textit{Chance} method generates a semantic scanpath by randomly selecting points from the image space. The \textit{Random} method generates semantic scanpaths by randomly selecting one scanpath of a human observer viewing a different scene. We obtained the predicted scanpaths of other methods by re-executing their source codes and generated corresponding semantic scanpaths according to the definition in Section \ref{sec.ssdef}.

Table \ref{T1} lists the performance of multiple methods under different evaluation metrics. First of all, all considered methods perform consistently when evaluated on existing metrics and new attention graph based metrics. This suggested the rationality of using attention graph as a representation of visual attention. However, we can see that the performance of human (i.e., the score by intra-observer comparison) is quite low with the ScanMatch and Sequence Score methods. This indicates that high intra-observer variability still exists with the coding of scanpath in these methods. In contrast, with the proposed attention graph and corresponding evaluation metrics, the evaluation with intra-observers obtains a high score. This indicates that the proposed attention graph represents the common patterns of human visual attention better. 

Moreover, some deep learning methods (e.g., VQA \citep{chen2021predicting}) seem to achieve human performance (e.g., 0.4118 vs. 0.4283 with Sequence Score) when evaluating with ScanMatch and Sequence Score methods. In contrast, all existing methods perform significantly worse than human observers with the proposed metrics based on attention graph. This reveals that existing models still have a significant gap in predicting human attention. Relatively,  IRL \citep{xia2019predicting} and VQA \citep{chen2021predicting} obtain good performance in semantic scanpath prediction. The IRL \citep{xia2019predicting} method may benefit from integrating global and local information in measuring center-surround relationships; while the  VQA \citep{chen2021predicting} method generates scanpaths by learning question-specific attention patterns in visual question answering task, which helps introduce general semantic information. Therefore, the attention graph is a better representation in evaluating attention-related prediction methods by embedding semantic information.

\begin{table*} 
  \centering
  \caption{The performances of existing scanpath prediction methods under different evaluation metrics (i.e., $S_{scan}$ in Eq.(\ref{E1})  and  $S'_{scan}$ in Eq.(\ref{E2})).}
  \label{T1}
  \begin{tabular}{c|c|c|c|c|c|c}
  \hline
  \multirow{2}{*}{Methods} & \multirow{2}{*}{ScanMatch} & \multirow{2}{*}{Sequence Score} & \multicolumn{2}{c|}{SemScan(obj)} & \multicolumn{2}{c} {SemScan(att)} \\
  \cline{4-7}
       &  &   &  $S_{scan}$ & $S'_{scan}$  & $S_{scan}$ & $S'_{scan}$ \\
  \hline
  Human & 0.3941 & 0.4283 & 0.6668 & 0.6754 & 0.7876 &0.7825 \\
  \hline
  Chance  & 0.1603 & 0.2701 & 0.3106 & 0.3055 & 0.3853 & 0.3521  \\
  Random & 0.1950 & 0.1929 & 0.2906 & 0.2763 & 0.3476 & 0.3495 \\
  \hline
  ITTI \citep{itti1998model} & 0.2041 & 0.2468 & 0.3206 & 0.3116 & 0.4022 & 0.3896 \\ 
  CLE \citep{boccignone2004modelling} & 0.2528   & 0.2982 & 0.4349 & 0.4307 & 0.5541 & 0.5388 \\
  SGC \citep{sun2014toward}  &  0.2383 &  0.2756 & 0.3874 & 0.3752 & 0.5019 & 0.4797  \\ 
  IRL \citep{xia2019predicting}   &  0.3135  &  0.3457 & 0.5638 & 0.5504 & 0.6780 & 0.6613 \\
  SaltiNet \citep{assens2017saltinet} & 0.1434   &  0.1996 & 0.3147 & 0.3151 & 0.4163 & 0.4090 \\
  PathGAN \citep{assens2018pathgan}&  0.0546 &  0.0678 & 0.1798 & 0.1760 &  0.2197 & 0.2150 \\
  VQA \citep{chen2021predicting} &  0.3773 &  0.4118 & 0.5657 & 0.5600 & 0.6966 & 0.6955 \\
  \hline
  \end{tabular}
\end{table*}

\begin{figure*}[tb]
\centering
\includegraphics[width=6.0in]{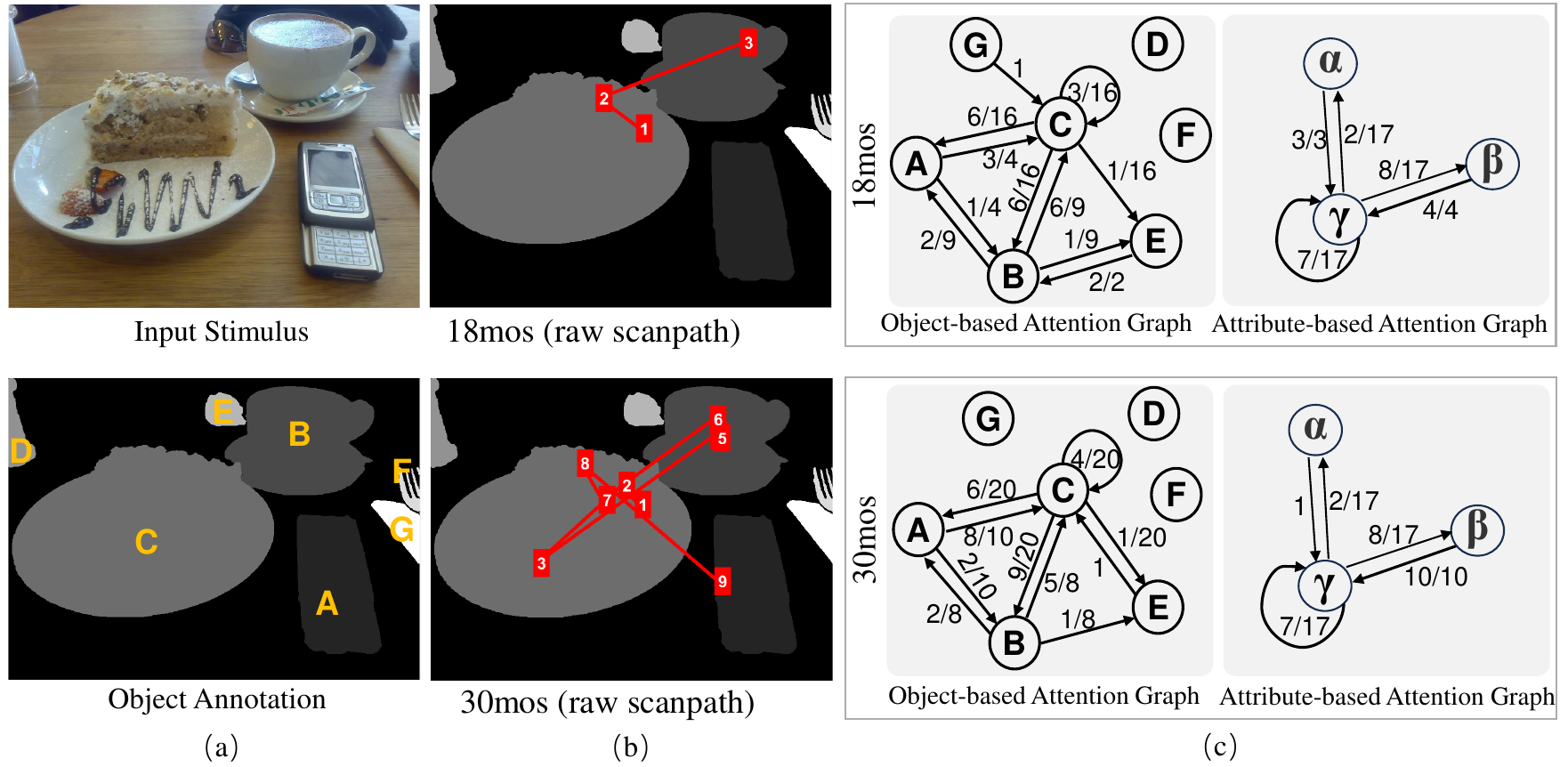}
\caption{Comparisons of the attention graphs between two age groups (18- and 30-month-old toddlers are denoted as 18mos and 30mos respectively). (a) the input stimulus and corresponding annotations, (b) examples of raw scanpath of individual 18- and 30-month-old toddlers, (c) attention graphs of 18- and 30-month-old toddlers. Note that the self-loop in the attention graph (e.g., node C) indicates there is one observer whose all fixations are located on a single object (i.e., no gaze shifting among objects).}
\label{FigAgeCom}
\end{figure*}

\begin{figure*}[tb]
\centering
\includegraphics[width=4.5in]{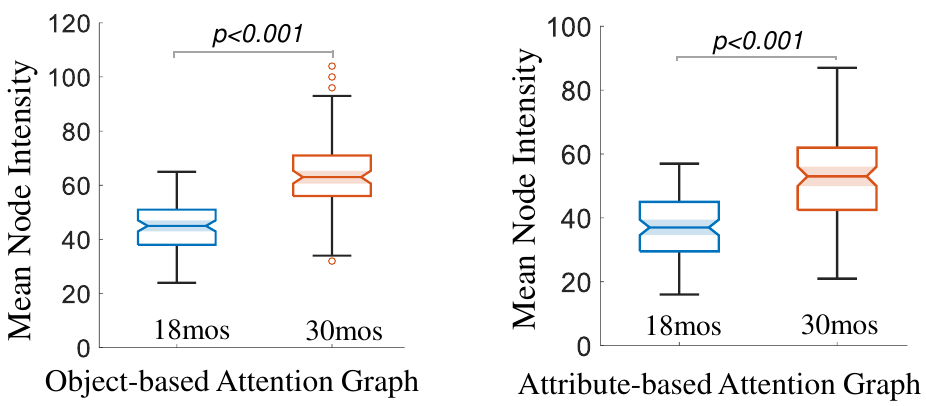}
\caption{Comparisons of the attention graphs between two age groups in the mean node intensity, where the node intensity means the total number of gaze shifts in each attention graph.}
\label{FigAgeMNI}
\end{figure*}

\subsection{Applications of Attention Graph}
To verify the potentials of the proposed attention graph on coding cognition information, subsequent tasks including age classification and autism spectrum disorder screening are considered to evaluate the semantic scanpath.

\subsubsection{Age Classification}
Firstly, we employ an age classification task to determine toddlers’ ages based on their visual attention recorded by eye-tracking. The basic logic of this task is that the differences in visual behaviors could reflect meaningful changes in toddlers cognitive processes when viewing scenes \citep{dalrymple2019machine}. Dalrymple et al. provided a  dataset that contains eye-tracking data from 18- or 30-month-old month toddlers (nineteen 18-month-olds and twenty-two 30-month-olds) \citep{dalrymple2019machine}. This dataset contains 100  images that are acquired from the OSIE dataset \citep{xu2014predicting}, and hence also contains the corresponding well-defined object and attributive annotations. Dalrymple et al. also built data-driven machine learning approaches to classify the age of toddlers based on eye tracking data and obtained high accuracy, which indicates the existence of clear difference in gaze patterns between different age groups. Therefore, we employ this dataset to evaluate the proposed attention graph and expect  to obtain high accuracy in distinguishing the age groups. 

Fig. \ref{FigAgeCom} presents examples of attention graph that show the different gaze patterns for different age groups. Fig. \ref{FigAgeCom}(a) lists the input stimulus and corresponding annotations. Two examples of individual raw scanpath for 18- and 30-month-old toddlers are shown in Fig. \ref{FigAgeCom}(b). The attention graph of 18- and 30-month-old toddlers (denoted as 18mos and 30mos in Fig. \ref{FigAgeCom}(c)) show differences in structures and connections. For example, the 18-month-old toddlers exhibit fewer gaze shifts but focus on more dispersed objects than the 30-month-old toddlers. In addition, statistical analysis based on all attention graphs quantitatively illustrates the distinguishability among different groups. Fig. \ref{FigAgeMNI} shows the significant difference between 18mos and 30mos groups in the mean node intensity (total number of gaze shifts) of attention graphs. 

\begin{figure*}[tb]
\centering
\includegraphics[width=5.0in]{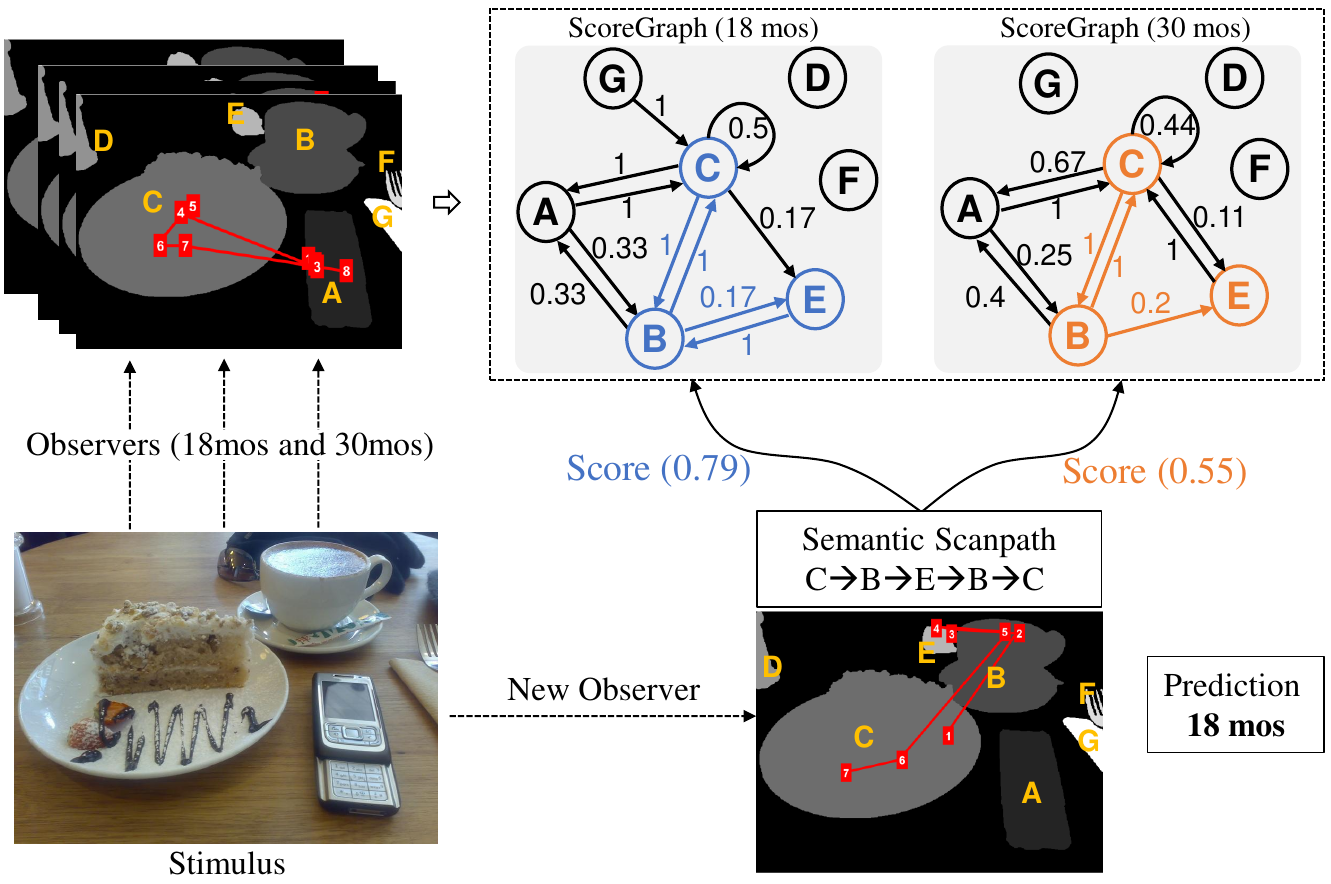}
\caption{Illustrating of classifying 18 or 30-month-old toddlers based on the gaze patterns represented by the attention graph. }
\label{FigAgeCls}
\end{figure*}

Subsequently, we conducted one more experiment using the leave-one-subject-out strategy to classify 18 or 30-month-old toddlers based on the gaze patterns represented by the attention graph. Fig. \ref{FigAgeCls} illustrates the general processing of age classification using attention graph on single scene. Specifically, we first built the attention graph for both groups of toddlers based on observer's eye-tacking data and object annotations. At the stage of prediction, one new observer's eye-tacking data is encoded into semantic scanpath and then simply computed the scores (i.e., $S_{scan}$ in Eq.(\ref{E1}) or $S'_{scan}$ in Eq.(\ref{E2})) with the attention graph of two groups respectively. The test samples are classified into the group with high score. Finally, the child is classified by voting based on the scores over all used stimuli (100 images in this study). That means the proposed method can classify 18 or 30-month-old toddlers directly based on the attention graph without any extra feature extraction and learning processing from stimuli and eye-tracking data.  

Table \ref{Tage} lists the performance of age classification. AttentionGraph(obj) and AttentionGraph(att) indicate the object-based and attribute-based attention graph respectively. The proposed method outperforms Dalrymple et al. (Baseline, with SVM classifier) but is slightly worse than Dalrymple et al. (with CNN classifier). The performance of our proposed method is 0.80 (33/41) and the performance Dalrymple et al. (CNN) \citep{dalrymple2019machine} is  0.83 (34/41). This accuracy indicates that only one more sample was misclassified by our proposed method. Dalrymple et al. built a deep learning model to extract high-level representations of an image and predict the difference between saliency maps of two group toddlers. Although the proposed method is slightly inferior to Dalrymple et al. (CNN) \citep{dalrymple2019machine}, we argue that the proposed method's distinctive advantage is no extra feature extraction processing for age classification. We can conclude that our attention graph is a simple and efficient representation in coding gaze patterns for different age groups, and contributes to comparable performance compared to the deep learning method proposed by Dalrymple et al. \citep{dalrymple2019machine}.

\begin{table*} 
  \centering
  \caption{The performance of age classification. The fractions in brackets indicate the number of correctly classified samples in total number of samples.}
  \label{Tage}
  \begin{tabular}{c|c}
  \hline
  Methods & Accuracy \\
  \hline
  Dalrymple et al. (Baseline) \citep{dalrymple2019machine} & 0.68 (28/41)  \\
  Dalrymple et al. (CNN) \citep{dalrymple2019machine} & \textbf{0.83 (34/41)}  \\
  \hline
 AttentionGraph(obj)+$S_{scan}$ & 0.71 (29/41) \\ 
 AttentionGraph(obj)+$S'_{scan}$ & 0.78 (32/41) \\
 AttentionGraph(att)+$S_{scan}$ & 0.80 (33/41) \\ 
 AttentionGraph(att)+$S'_{scan}$ &  0.80 (33/41)  \\
  \hline
  \end{tabular}
\end{table*}

\subsubsection{Autism Spectrum Disorder Screening} 
In addition, numerous recent studies have shown that gaze patterns can be useful in screening Autism Spectrum Disorder (ASD) \citep{chen2019attention, wang2015atypical, gutierrez2021saliency4asd}. Therefore, we also employed the ASD screening task to investigate the contribution of the proposed semantic scanpath in this healthcare societal challenge. The Saliency4ASD dataset \citep{gutierrez2021saliency4asd} provides 300 images and eye-tracking data of children with ASD and with Typical Development (TD) to support the research on visual attention modelling towards ASD screening. Specifically, Guti{\'e}rrez et al. collected eye movement data from 14 children with ASD and 14 children with TD, all of them have ages between 5 and 12 years with an average of 8 years. However, the Saliency4ASD dataset does not provide subject IDs for each group, therefore the $i^{th}$ fixation sequences from the same group is regarded as the same subject, maintaining consistency with previous work by Chen et al. \citep{chen2019attention}. Moreover, this dataset does not provide the corresponding object and semantic annotations that are needed for building a semantic scanpath. 

\begin{figure*}[tb]
\centering
\includegraphics[width=6.0in]{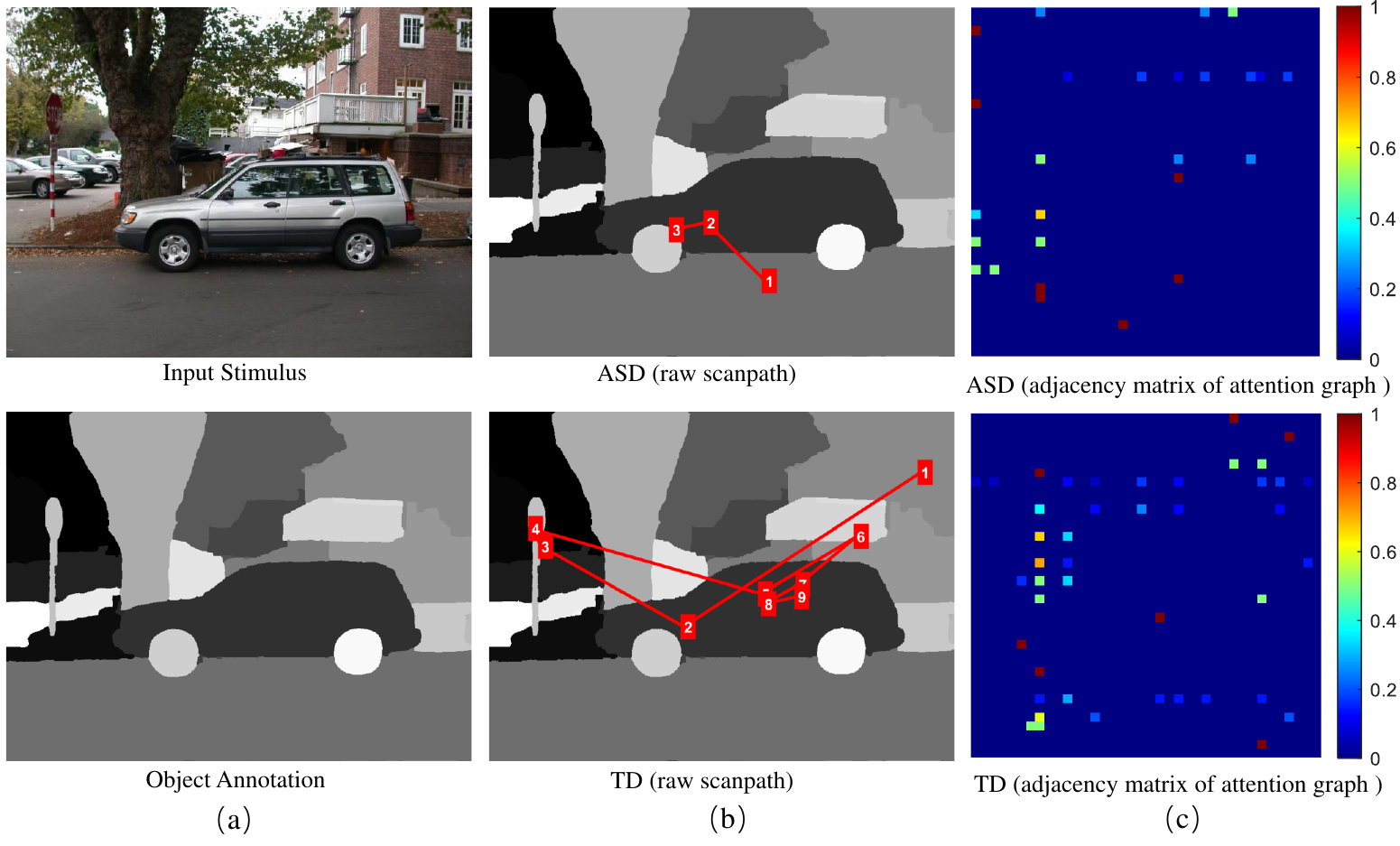}
\caption{Comparisons of the attention graphs between ASD and TD groups. (a)the input stimulus and corresponding annotation, (b) examples of raw scanpath of individual ASD and TD children, (c) adjacency matrices of attention graphs for ASD and TD children. Note that the adjacency matrix is used to show the difference of attention graphs between ASD and TD groups, as the attention graph of scenes in the Saliency4ASD dataset is too complex to visualize in graphs.}
\label{FigASDCom}
\end{figure*}

In this study, we first annotated objects in the Saliency4ASD dataset. Specifically, to simplify the annotation process, we employed the SAM model \citep{kirillov2023segment} to initially segment the images and then fine-annotated the main objects by hand. It should be noted that we did not annotate semantic attributes of each object or region. This is because the scenes in the Saliency4ASD dataset are quite complex (see Fig. \ref{FigASDCom} for an example), e.g., there are quite more attributive classes and crowding small objects in outdoor scenes, which significantly increases the difficulty of annotation. 

\begin{figure}[tb]
\centering
\includegraphics[width=2.0in]{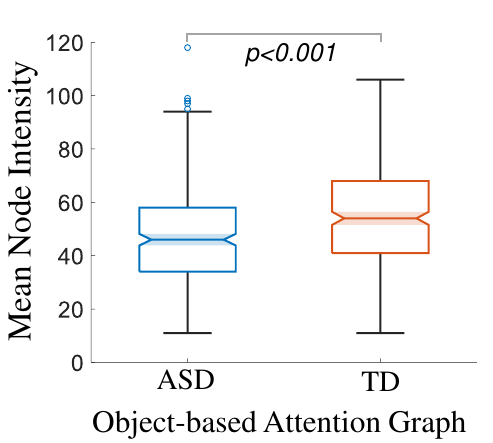}
\caption{Comparisons of the attention graphs between ASD and TD groups in the mean node intensity, where the node intensity means the total number of gaze shifts in each attention graph.}
\label{FigASDMNI}
\end{figure}

Fig. \ref{FigASDCom} also presents examples of attention graph that show the different gaze patterns between ASD and TD groups. The attention graph of ASD and TD groups show differences in structures and connections in corresponding adjacency matrices. Note that we use the adjacency matrix to show the difference of attention graphs between ASD and TD groups in Fig. \ref{FigASDCom}, as the attention graph of scenes in the Saliency4ASD dataset is too complex to visualize in graphs. Moreover, Fig. \ref{FigASDMNI} shows the significant difference between ASD and TD groups in the mean node intensity of attention graphs. 

To screen children with ASD, we conducted experiments for ASD and TD classification task using the leave-one-subject-out strategy, as per previous works \citep{chen2019attention}.  Table \ref{Tasd} lists the performances compared to other methods, e.g., CNN in \citep{chen2019attention}. Similarly, our method based on the $AttentionGraph(obj)+S_{scan}$, without extra feature learning processing, obtains the same performance compared to the deep learning method proposed by Chen et al. \citep{chen2019attention}. 

\begin{table*} 
  \centering
  \caption{The performance of ASD screening. The fractions in brackets indicate the number of correctly classified samples in total number of samples.}
  \label{Tasd}
  \begin{tabular}{c|c}
  \hline
  Methods & Accuracy \\
  \hline
  Chen et al. (Independent) \citep{chen2019attention} & 0.89 (25/28) \\
  Chen et al. (Full) \citep{chen2019attention} & \textbf{0.93 (26/28)} \\
  \hline
 AttentionGraph(obj)+$S_{scan}$ & \textbf{0.93 (26/28)}   \\ 
 AttentionGraph(obj)+$S'_{scan}$ & 0.86 (24/28) \\
  \hline
  \end{tabular}
\end{table*}

In summary, the experimental results mentioned above demonstrate that age classification and ASD screening can be effectively achieved using the proposed attention graph representation without extra feature learning processing. Although the performance is slightly inferior (Table \ref{Tage}) or equivalent (Table \ref{Tasd}) to existing methods, the attention graph only relies on object-based attention rather than pixel-wise eye-tracking data. This implies that there are potential low-cost technologies (e.g., webcam-based eye tracking) to obtain attention graphs instead of using high-cost eye trackers. Consequently, the proposed method offers distinct advantages for developing low-cost technologies applicable in real  environments, such as ASD screening.

\section{Conclusion and Discussion}
This paper proposes a new representation of visual attention called attention graph and builds an evaluation benchmark based on the attention graph. Our attention graph can better reveal the intrinsic common patterns of visual fixations and gaze-shift behaviors of human observers when viewing a visual scene. The proposed method and metrics can be directly used to analyze human's visual scanpaths, and then contribute to downstream tasks like ASD screening. Experiments on semantic scanpath evaluation show the significant contribution of the proposed method. We expect the proposed attention graph to provide an efficient tool to investigate cognition-related processing with visual attention behaviors and also contribute to advancing attention-based computer vision methods. 

As a newly defined representation of visual attention, there are several open questions that require further study. For instance, we need to explore how to develop an efficient models to generate attention graph from visual scenes and corresponding evaluation methods. Additionally, investigating the applications of attention graphs in active vision and medical diagnosis will be crucial research directions.

In addition, considering that the attention graph is based on semantic annotations, varying scaled or more detailed object and attribute annotations may result in different coding of attention graphs. Meanwhile, inaccurate semantic annotation, e.g., missing small objects, incomplete annotation, or over-segmentation, may also affect the coding of attention graphs. However, in this study, we aim to build a generic representation of attention graph by utilizing the annotations provided in the OSIE dataset\citep{xu2014predicting}, which can be easily refined when new well-annotated datasets will be available in the future.

\backmatter

\bmhead{Acknowledgements}
This study was supported by the STI2030-Major Projects (2022ZD0204600) and the National Natural Science Foundation of China (62076055, 62476050). This work was also partly supported by Sichuan Science and Technology Program (2024NSFSC0657), Huzhou Science and Technology Program (2023GZ13) and the Fundamental Research Funds for the Central Universities (Y03023206100215).

\section*{Declarations}
\bmhead{Data Availability Statement}
The authors confirm that the data supporting the findings of this study are openly available at https://github.com/NUS-VIP/predicting-human-gaze-beyond-pixels for OSIE dataset, at https://osf.io/v5w8j/ for age classification dataset, at https://saliency4asd.ls2n.fr for Saliency4ASD.  Our codes for data analysis are available upon reasonable request.


\bibliographystyle{sn-basic}
\bibliography{Refs}

\end{document}